# Causal Mechanism-based Model Constructions

Tsai-Ching Lu & Marek J. Druzdzel
Decision Systems Laboratory
School of Information Sciences
and Intelligent Systems Program
University of Pittsburgh
Pittsburgh, PA 15260
{ching,marek}@sis.pitt.edu

Tze Yun Leong
Medical Computing Laboratory
Department of Computer Science
School of Computing
National University of Singapore
Singapore 119260
leongty@comp.nus.edu.sg

## Abstract

We propose a framework for building graphical causal model that is based on the concept of causal mechanisms. Causal models are intuitive for human users and, more importantly, support the prediction of the effect of manipulation. We describe an implementation of the proposed framework as an interactive model construction module, *ImaGeNIe*, in *SMILE* (Structural Modeling, Inference, and Learning Engine) and in *GeNIe* (SMILE's Windows user interface).

## 1 INTRODUCTION

Graphical probabilistic models, such as Bayesian networks and influence diagrams, have become popular modeling tools for supporting decision making under uncertainty. The normative character of the graphical decision models guarantees the correctness of the inference procedure. Consequently, the quality of the advice suggested by the models depends directly on the requisiteness of the models. A model is *requisite* if it contains everything that is essential for solving the problem and no new insights about the problem will emerge by elaborating on it (Philips 1982). To build a requisite model requires human intuition and creativity since the notion of requisiteness is subjective. Construction of graphical models, therefore, is laborious and demanding in terms of domain expertise. While support for obtaining model parameters, such as prior and conditional probability distribution, has received much attention in behavioral decision theory literature (see von Winterfeldt and Edwards (1988) for a review) and in artificial intelligence (Druzdzel & van der Gaag 2000), relatively little work has been done on composing model structure. At the same time, there are strong indications that the quality of advice is more sensitive to the model structure than to the precision of its numerical parameters (Pradhan *et al.* 1996).

There are essentially four approaches to aid model building. The first approach focuses on providing more expressive building tools. The Noisy-OR model (Pearl 1988; Henrion 1989) and its generalizations (Díez 1993; Srinivas 1993) simplify the representation and elicitation of independence interactions among multiple causes. Heckerman (1990) developed the *similarity network* and *partition* as tools for representing *subset independence* to facilitate the structure construction and probability elicitation. The second approach, usually referred to knowledge-based model construction (KBMC), emphasizes aiding model building by automated generation of decision models from a domain knowledge-base guided by the problem description and observed information (see a special issue at the journal IEEE Transactions on Systems, Man and Cybernetics on the topic of KBMC (Breese, Goldman, & Wellman 1994)). The third approach focuses on algorithms that can learn the model structure and parameters from a database of observations (Cooper & Herskovits 1991; Pearl & Verma 1991; Spirtes, Glymour, & Scheines 1993). Although model construction from data can reduce the knowledge engineering effort, the learning approach faces other problems such as small data sets, unmeasured variables, missing data, selection bias, and the flexibility of model granularity.

While we acknowledge that in the future it may be possible to build powerful computer systems that will model human creativity, sense for relevance, and simplicity, we believe that these tasks are and will long be performed better by humans. Our view is that model building, a task that relies on all these capacities, is best implemented as an interactive process. The fourth approach on aiding model construction that is most related to our work is to apply system engineering and knowledge engineering techniques for aiding the process of building Bayesian networks. Laskey and Mahoney (1996; 1997) address the issues of modularization, object-orientation, knowledge-base, and evalua-tion in a spiral model of development cycle. Koller and Pfeffer (1997; 1999) developed Object-Oriented Bayesian Networks (OOBN) that use objects as organizational units to reduce the complexity of modeling and increase the speed of inference.

Our approach on aiding model construction is based on the concept of causal mechanisms. Causal mechanisms, which are local interactions among domain variables, are building blocks that determine the causal structure of a model. As they encode our understanding of local interactions and are fairly model independent, causal mechanisms can be easily reused in various models. When the algebraic form of the interaction is known, causal mechanisms are captured by so called *structural equations*. When less information is available about the interaction, it can be specified in a probabilistic format. As shown by Druzdzel and Simon (1993), conditional probability tables in Bayesian networks that model causal relations among their variables can be also viewed as descriptions of causal mechanisms. Similarly to object-hierarchy abstraction, causal mechanism can be organized hierarchically in nearly decomposable system (Iwasaki & Simon 1994). At the same time they provide a valuable heuristic for acquiring and managing knowledge: causality.

In our framework, we encode causal mechanisms as functional relations among variables and, wherever causal mechanisms are asymmetric, the direction of causal influence among variables. We extend Simon's causal ordering algorithm (Simon 1953) to develop a modeling process that uses the output graph of this algorithm in the interaction with users. We assist the model building process by helping user (1) to identify a set of mechanisms related to the current model and to bring them into model workspace (2) to integrate the newly added mechanisms with the model under construction (3) to specify the variables that can be manipulated, and (4) to extract reusable causal mechanisms from existing models into the knowledge base. The final model structures generated by our modeling process are guaranteed to be causal if the underlying structural equations reflect causal mechanisms of the modeled problem.

In addition to being intuitive for human users and facilitating crucial user interface functions such as explanation, causal models support prediction of the effect of manipulation, i.e., changes in structure (Simon 1953; Spirtes, Glymour, & Scheines 1993; Pearl 1995). The users of such models (and that includes autonomous robots) can ask questions like "What will happen if I perform action $A$?" Manipulation is especially important in strategic planning, where it is important to derive creative decision options and not only evaluate existing decision options. In the process of creating a model, a user may want to explore the possibility of manipulating its different elements. Supporting this manipulation is not straightforward, as some mechanisms may be reversible, i.e., acting in reverse direction. For example, when driving up the hill, car engine causes the wheels to turn; but when driving down the hill in a low gear, the model should be able to predict that the wheels will cause the engine to slow down. Our approach supports causal modeling that includes reversible causal mechanisms and offers an integrated framework for building and using causal models.

The remainder of this paper is structured as follows. Section 2 gives an overview of structural equation models, causal mechanisms, and how these support changes in structure. Section 3 discusses the process of interactive model construction, including issues related to the representation of causal mechanism, assistant interface, and the extension of causal ordering algorithm. Section 4 presents an example of user interaction with our system, *ImaGeNIe*. Finally, we discuss the implications of our approach and outline the direction for our future work.

## 2 STRUCTURAL EQUATION MODELS

When scientists study phenomena or problems, they normally focus on *systems*, pieces of the real world that can reasonably be studied in isolation. Scientists identify the relevant variables, the ranges of the variables' values, and the relations among variables to form abstractions of these systems, known as *models*. One way of representing models is by systems of structural equations where each structural equation describes a conceptually distinct causal mechanism active in the system. Such systems are known as *Structural Equation Models* (SEMs) (Haavelmo 1943; Simon 1953). A *structural equation* describing a causal mechanism $\mathcal{M}$ is often encoded as an *implicit function*

$$f_\mathcal{M}(V_1, V_2, V_3, \ldots, V_n),$$

where $f$ is some algebraic function and its arguments $V_i$ are variables that directly participate in the mechanism $\mathcal{M}$.

A variable in a SEM is *exogenous* if it summarizes an outside influence on the system, i.e., its value is determined outside of the model. An exogenous variable is *truly exogenous* if it represents a variable in the real world system that we cannot manipulate without changing the boundaries of the system. An exogenous variable is a *policy variable* if it represents a variable that we can manipulate, i.e., set its value. For example, we normally model outside temperature as a truly exogenous variable in an agricultural model, but we can model the temperature as a policy variable in a model of a greenhouse. For each exogenous variable, there is a *value assignment structural equation* to designate the observed value (or a probability distribution over observed values) for the truly exogenous variable or the chosen value for the policy variable. A variable in a SEM is *endogenous* if its value is derived by substituting the values of exogenous variables into the *core structural equations* that depict the relations among modeled variables in the system and by solving these equations in SEM.

A SEM $\mathcal{S}$ with $m$ causal mechanisms and $n$ variables is represented as

$$\mathcal{S} = \bigcup_{i=1}^{m} f_{\mathcal{M}_i}(V_1, V_2, V_3, \ldots, V_n).$$

Since the knowledge of which variables participate in which mechanisms is sufficient to determine the direction of causation,[1] in the remainder of this paper we will only use *structure matrix* (Druzdzel & Simon 1993), a qualitative representation of a SEM.

**Definition 1** *(structure matrix)* A structure matrix $\mathcal{A}$ of a SEM $\mathcal{S} = \bigcup_{i=1}^{m} f_{\mathcal{M}_i}(V_1, V_2, V_3, \ldots, V_n) = 0$ is a $m \times n$ matrix with element $a_{ij} = $ x if $V_j$ participates in $f_{\mathcal{M}_i}$, where x is a marker, and $a_{ij} = 0$ otherwise.

Let $\mathcal{A}_{m \times n}$ be the structure matrix of a SEM $\mathcal{S}$ with $m$ equations and $n$ variables. $\mathcal{S}$ is *non-over-constrained* if following property holds.

**Definition 2** *(non-over-constrained system)* A system of $m$ structural equations $\mathcal{S}$ is *non-over-constrained* if in any subset of $k \leq m$ equations of $\mathcal{S}$ at least $k$ different variables appear with nonzero coefficients.

A non-over-constrained $\mathcal{A}_{m \times n}$ is *self-contained* if $m = n$. A non-over-constrained $\mathcal{A}_{m \times n}$ is *under-constrained* if $m < n$. $\mathcal{A}_{m \times n}$ is *over-constrained* if it violates non-over-constrained property.

**Example:** The University Performance Budget Planning Model (UPBPM) (Simon, Kalagnanam, & Druzdzel 2000) is comprised of 38 core equations that describe interactions among 88 variables in the university strategic budget planning context. The model has been adopted by the Office for Planning and Budget at Carnegie Mellon University for the purpose of strategic planning of university operations.

The following simple model, *StudentFacultyRatio* model, extracted from UPBPM, consists of one core equations and two value assignment equations and describes the interaction among three variables: *StudentFacultyRatio (SFR)*, *NumberOfStudents (NS)*, and *NumberOfFaculty (NF)*.

---

[1] Only when calculating the strength of the influences, we need the exact form of equations.

The corresponding structure matrix for this self-contained model is shown at the right hand side.

$$\left\{ \begin{array}{lll} f_1 : & NS & = 22102 \\ f_2 : & NF & = 3006 \\ f_3 : & SFR & = NS/NF \end{array} \right. \qquad \begin{array}{c|ccc} & NS & NF & SFR \\ \hline f_1 & \text{x} & 0 & 0 \\ f_2 & 0 & \text{x} & 0 \\ f_3 & \text{x} & \text{x} & \text{x} \end{array}$$

□

### 2.1 Causal Ordering

As shown by Simon (1953), a self-contained SEM exhibits asymmetries that can be represented by a directed acyclic graph and interpreted causally. Simon developed a causal ordering algorithm that takes a self-contained structure matrix $\mathcal{A}$ as input and outputs a *causal graph* $G = \{N(G), A(G)\}$, where the nodes, $N(G)$, are sets of variables and the arcs, $A(G)$, describe causal relations among them.

Let $\mathcal{B}$ be a subset of equations in a non-over-constrained SEM and $\mathcal{C}_{p \times q}$ be the structure matrix of $\mathcal{B}$. We say that $\mathcal{B}$ is a *self-contained subset* if $p = q$; $\mathcal{B}$ is a *under-constrained subset* if $p < q$. A self-contained subset is *minimal* if it does not contain any self-contained (proper) subsets itself. A minimal self-contained subset is a *strongly coupled component* if it contains more than one equation, which usually represents a feedback system in the real world.

The causal ordering algorithm starts with *identifying* the minimal self-contained subsets in input $\mathcal{A}$. These identified minimal self-contained subsets are called *complete subsets of 0-th order* and a node is created for each subset. Next, the algorithm removes the equations of the complete subsets of 0-th order from $\mathcal{A}$ as *solving* the values of variables. Then it removes all variables that occur in the complete subsets of 0-th order from the remaining equations in $\mathcal{A}$ as *substituting* the values of solved variables into remaining equations. The remaining set of equations is called the *derived system of first order*, a self-contained structure. The algorithm repeats the process of identifying, solving, and substituting on the derived system of $k$-th order until it is empty. In addition, whenever a node $m$ is created for a minimal self-contained subset $\mathcal{M}$, the algorithm refers the set of equations $E_{\mathcal{M}}$ of $\mathcal{M}$ back to the original set of equations $OE_{\mathcal{M}}$ in $\mathcal{A}$ and adds arcs from the nodes representing variables in $OV_{\mathcal{M}} \setminus V_{\mathcal{M}}$ to $m$, where $V_{\mathcal{M}}$ is the set of variables participating in $E_{\mathcal{M}}$ and $OV_{\mathcal{M}}$ is the set of variables participating in $OE_{\mathcal{M}}$.

**Example:** The UPBPM (Simon, Kalagnanam, & Druzdzel 2000) implements Simon (1953) causal ordering algorithm that given an assignment of values to 50 exogenous variables, derives the structure of the model.

When applying the causal ordering algorithm to the structure matrix of *StudentFacultyRatio* model, we first identify

$f_1$ and $f_2$ as the complete subset of 0-th order. After solving and substituting of $NS$ and $NF$, we then identify $f'_3$ as the complete subset of 1-st order. The structure matrix and the corresponding causal graph are shown below.

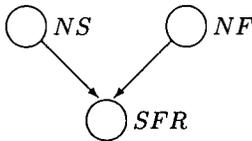

|       | $NS$ | $NF$ | $SFR$ |
|-------|------|------|-------|
| $f_1$ | x    | 0    | 0     |
| $f_2$ | 0    | x    | 0     |
| $f_3$ | x    | x    | x     |

Given the causal graph, we can read off the causal relations among the nodes by focusing on the node of interest and its parents. For example, $SFR$ directly depends on $NF$ and $NS$. □

Notice that the causality that we read off causal graphs is defined within models and causal asymmetries arise when mechanisms are placed in context. If the context has changed, it may result in changes in structure.

## 2.2 Changes in Structure

The main value of structural equation models is that they support prediction of the effects of changes in structure, i.e., external manipulations that intervene in the mechanisms captured by the original system of equations. Such changes are modeled by modifying the equations that describe the affected mechanisms and leaving those equations that correspond to unaffected mechanisms unmodified. The causal ordering algorithm applied to the modified SEMs derives the new causal structure of the system.

Normally, the effect of external manipulation is local and, when related back to the graph, amounts to arc cutting (Pearl 1995; Spirtes, Glymour, & Scheines 1993). The assumption underlying the arc-cutting operation is that imposing a value on a variable by an external intervention makes that variable independent of its direct causes. This assumption is valid for mechanisms with strong asymmetric relationship between a variable and its causes; for example, wearing sunglasses protects our eyes from the sun but it does not make the sun go away. However, when a model contains *reversible causal mechanisms* (Simon 1953; Druzdzel & van Leijen 2000), manipulation can have a drastic effect on the graph.

**Example:** From the causal graph of *StudentFacultyRatio* model in previous example, we know that changing $NS$ will affect $SFR$ but not $NF$. Now, consider that the budget planning officer would like to set the *StudentFacultyRatio* to advertise their faculty availability. If needed, she is willing to adjust the *NumberOfFaculty* (e.g., hire more faculty). According to the revised modeling context, she needs to designate the variable $SFR$ as exogenous, e.g., $f_4 : SFR = 10$, and release $f_2 : NF = 3,006$. The resulting structure matrix and corresponding causal graph are:

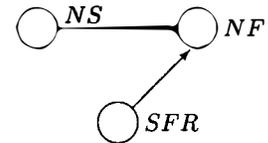

|       | $NS$ | $NF$ | $SFR$ |
|-------|------|------|-------|
| $f_1$ | x    | 0    | 0     |
| $f_3$ | x    | x    | x     |
| $f_4$ | 0    | 0    | x     |

In the revised system, the causal ordering shows that $NF$ has become an endogenous variable affected by $NS$ and $SFR$. Now, changing the number of students will affect the number of faculty. Manipulation has lead to a change in structure. □

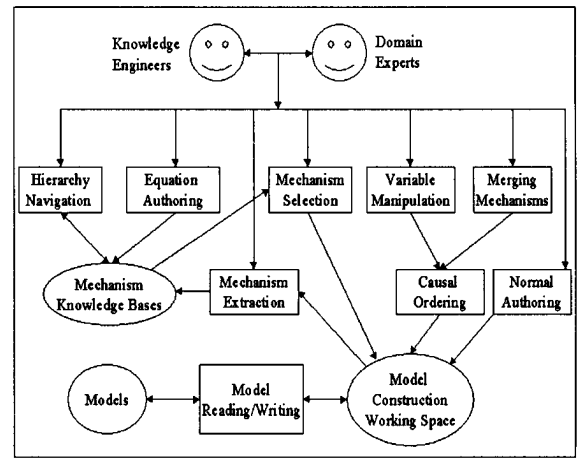

Figure 1: Interactive and Iterative Model Construction System Architecture. The arcs show the direction of the information flow.

## 3 INTERACTIVE MODEL CONSTRUCTION

We have developed an interactive and iterative model construction environment, *ImaGeNIe*, that assists users in building graphical decision model in causal form. We use the causal ordering algorithm to generate the causal model structures which can later be associated with different node types and parameters and transformed into Bayesian networks or influence diagrams. Figure 1 shows the architecture of *ImaGeNIe*. It includes three knowledge structures: *mechanism knowledge bases*, which hold domain knowledge expressed as causal mechanisms, *model building workspace*, which serve as a blackboard for model composition, and *models*. The domain knowledge can be maintained either by the *equation authoring interface*, where model builders can compose structural equations directly, or by the *mechanism extraction operation* that enables model builders to extract reusable causal mechanisms from existing models. Model builders can use *hierarchy navigation interface* to locate the mechanisms of interest and select them into the model building workspace with assistance of the *mechanism selection operation*. In addition to mechanism selection and traditional model authoring operations, model builders can *manipulate variables* and *merge mechanisms* as model building process evolves. The underlying *causal ordering* module will restructure the models according to the user actions.

### 3.1 Knowledge Representation

In *ImaGeNIe*, the fundamental knowledge representation units are causal mechanisms, which are encoded as structural equations. For example, we can specify the student faculty ratio as $f_3(SFR, NS, NF)$. Users may optionally provide explicit functions for causal mechanisms such as algebraic functions, conditional probability tables, truth tables, value/utility tables, and choice tables.

While most mechanisms will be described in one, perhaps their only, mode of operation, some mechanisms are reversible in the sense of being flexible as to the direction of causality that they imply when they are embedded in different contexts. We define the *manipulativeness* and *observability* for each variable in our domain knowledge base to express the characteristics of the variable that may aid in the process of model building. Along with the manipulativeness characteristic, a variable can be *truly exogenous*, *manipulatable*, or *truly endogenous*. For the sun and sunglasses example, we may use two structural equations $f_5(S, G)$ and $f_6(S)$ to describe causal relation between $S$ and $G$ and assign $S$ as a *truly exogenous* variable to express the fact that it is impossible to manipulate the sun in the current modeling domain. We may assign $G$ as *manipulatable* to designate it as a *potential* policy variable. A variable is *truly endogenous* if its value has to be derived from embedded mechanisms. The observability is important in deciding whether adding this variable (observable or unobservable) will be of benefit to the model. In the diagnostic domain, it may be desired to develop the cost model that can associate *manipulation cost/observation cost* with manipulatable/observable variables.

Our domain knowledge base is organized as a hierarchical system that consists of subsystems and causal mechanisms as its fundamental building elements. The hierarchical approach not only helps domain experts to express their domain knowledge in cognitively meaningful units but also helps knowledge engineers to access stored mechanisms easily. Our approach is similar to type-hierarchy in (Koller & Pfeffer 1997; Laskey & Mahoney 1997) but without imposing the inheritance constraint since knowledge can be possibly organized hierarchically from different perspectives. More details on the syntax of our knowledge representation language can be found in (Lu 1999).

### 3.2 Extending Causal Ordering to Under-constrained Model

In *ImaGeNIe*, the model construction process is a reflection of our problem solving. The under-constrained models evolved in such process reveal different problem recognition stages. In an under-constrained model, the mechanisms are our observations of how the problem should be described so far. Model building process is strongly related to causal manipulation. The exogenous variables are those outside influences that have been committed. An under-constrained model cannot be drawn as a directed acyclic graph, as the direction of causal interactions is not completely determined until the model is self-contained. However, it is desired to have a graphical representation of under-constrained models during the whole process of model construction, since the graphical representation may help model builder identify her focus and change her commitments of the outside influences. We extend Simon's causal ordering algorithm to explicate the causal ordering that has been identified in under-constrained models. We also propose a graphical representation to depict the causal ordering results in an informative graphical form that aims to help user in model building.

In order to formalize our extensions, we need to restate the theorem that was originally proved by Simon (1953).

**Theorem 1** *Let $\mathcal{A}$ and $\mathcal{B}$ be two minimal self-contained subsets of equations of a non-over-constrained SEM, $\mathcal{S}$. Then the structural equations of $\mathcal{A}$ and $\mathcal{B}$, and likewise the variables in $\mathcal{A}$ and $\mathcal{B}$ are disjunct.*

Consider any subset $\mathcal{B}$ of the equations of a non-over-constrained SEM. We will denote the number of equations in $\mathcal{B}$ as $ne_\mathcal{B}$, and the number of variables appearing in $\mathcal{B}$ as $nv_\mathcal{B}$.

**Theorem 2** *Let $\mathcal{S}$ be a non-over-constrained system and $\mathcal{D}$ be the derived system of structural equations from $\mathcal{S}$ by applying identification, solving, and substitution. If $\mathcal{D}$ is not empty, then $\mathcal{D}$ is non-over-constrained.*

**Proof:** In the process of identification, let $\mathcal{M}$ be the union of all the minimal self-contained subsets, $\mathcal{M} = \mathcal{M}_1 \cup \mathcal{M}_2 \cup \ldots \cup \mathcal{M}_k$, and the remainder $\mathcal{R}$. We know $\mathcal{R}$ is not empty since $\mathcal{D}$ is not empty.

Suppose that $\mathcal{D}$ violates the non-over-constrained property. Then there exists a subset $\mathcal{E}'$ of $\mathcal{D}$ such that $ne_{\mathcal{E}'} > nv_{\mathcal{E}'}$.

Let $\mathcal{E}$ be the subset of $\mathcal{R}$ that $\mathcal{E}'$ derives from. We know that $ne_{\mathcal{E}} = ne_{\mathcal{E}'}$. Now, consider the subset $\mathcal{F} = \mathcal{M} \cup \mathcal{E}$. The equations of $\mathcal{M}$ and $\mathcal{E}$ are disjunct because $\mathcal{M}$ and $\mathcal{R}$ are disjunct and $\mathcal{E} \subseteq \mathcal{R}$. Therefore, $ne_{\mathcal{F}} = ne_{\mathcal{M}} + ne_{\mathcal{E}} = ne_{\mathcal{M}} + ne_{\mathcal{E}'}$. Since $\mathcal{E}'$ derives from $\mathcal{E}$ by substitution, the variables appearing in $\mathcal{E}$ are either in $\mathcal{M}$ or in $\mathcal{E}'$. Consequently, the variables in $\mathcal{F}$ are either in $\mathcal{M}$ or in $\mathcal{E}'$. Moreover, the variables in $\mathcal{M}$ and $\mathcal{E}'$ are disjunct because $\mathcal{E}'$ derives from $\mathcal{E}$ by substituting out the variables in $\mathcal{M}$. Therefore, $nv_{\mathcal{F}} = nv_{\mathcal{M}} + nv_{\mathcal{E}'}$. Since the equations of $\mathcal{M}_i$, and likewise the variables in $\mathcal{M}_i$, are disjunct by Theorem 1, we have $nv_{\mathcal{M}} = \sum nv_{\mathcal{M}_i}$ and $ne_{\mathcal{M}} = \sum ne_{\mathcal{M}_i}$. Hence $nv_{\mathcal{M}} = ne_{\mathcal{M}}$. Therefore, $ne_{\mathcal{F}} = ne_{\mathcal{M}} + ne_{\mathcal{E}} = ne_{\mathcal{M}} + ne_{\mathcal{E}'} > ne_{\mathcal{M}} + nv_{\mathcal{E}'} = nv_{\mathcal{M}} + nv_{\mathcal{E}'} = nv_{\mathcal{F}}$, i.e., the number of equations of $\mathcal{F}$ is greater than the number of variables of $\mathcal{F}$. In other words, the set $\mathcal{F}$ violates Definition 2 contradicting the fact that $\mathcal{S}$ is non-over-constrained. We conclude that $\mathcal{D}$ must be non-over-constrained. □

Given Theorem 2, we can keep applying identification, solving, and substitution operations on derived non-over-constrained system until either $\mathcal{D}$ is empty or there are no more minimal self-contained subsets that can be identified. If $\mathcal{D}$ is empty, we know that $\mathcal{S}$ is self-contained. If $\mathcal{D}$ is not empty and no more self-contained subsets can be identified, we know that $\mathcal{S}$ is under-constrained and we call $\mathcal{D}$ the *derived strictly under-constrained subsets*.

**Definition 3** *(strictly under-constrained subsets)* The strictly under-constrained subsets of a non-over-constrained SEM are those under-constrained subsets that do not contain any self-contained subsets.

**Theorem 3** A SEM, $\mathcal{S}$, is under-constrained if and only if there exists a derived strictly under-constrained subset in $\mathcal{S}$.

**Proof (sketch):** We can prove $\Rightarrow$ by construction and $\Leftarrow$ by contradiction given Theorem 2. See (Lu 1999) for the formal proof. □

Figure 2 outlines our extended causal ordering algorithm that is based on Theorem 3. The input of the algorithm is a non-over-constrained structure matrix $\mathcal{A}$. The output is a graph $G = \{\mathbf{V}, A(G)\}$, where the nodes $\mathbf{V}$ are variables and $A(G)$ is a set of directed, bi-directed, or undirected arcs. The algorithm essentially follows the steps of identification, solving, and substitution as Simon's causal ordering algorithm until there are no more self-contained subsets that can be identified from the derived system. The algorithm will explicitly depict the causal relations and relevant relations encoded in the strictly under-constrained subset, if there remains one.

The graph generated by our extended causal ordering algorithm is specifically designed to aid the process of

---

**Procedure ExtendedCausalOrdering**

**Input:** A structure matrix $\mathcal{A}$ of a non-over-constrained SEM.

**Output:** A graph $G = \{\mathbf{V}, A(G)\}$, where $\mathbf{V}$ are the variables in $\mathcal{A}$ and $A(G)$ is a set of directed, bi-directed, or undirected arcs.

Let $i := 0$ and $\mathcal{D}^0 := \mathcal{A}$
**while** there exists $\mathcal{M}^i \subseteq \mathcal{D}^i$, where $\mathcal{M}^i$ is the complete subset of $i$-th order and $\mathcal{D}^i$ is the derived structure of $i$-th order.

1. **for each** minimal self-contained subset $\mathcal{M}^i_j \in \mathcal{M}^i$, where $1 \leq j \leq |\mathcal{M}^i|$
   (a) Create nodes $N^i_j$ for all variables in $V_{\mathcal{M}^i_j}$.
   (b) Add arcs from the nodes represent $OV_{\mathcal{M}^i_j} \setminus V_{\mathcal{M}^i_j}$ to nodes in $N^i_j$, where $OV_{\mathcal{M}^i_j}$ is the set of the variables in $OE_{\mathcal{M}^i_j}$, the original equations of equations $E_{\mathcal{M}^i_j}$ of $\mathcal{M}^i_j$ in $\mathcal{A}$.
   (c) **if** $|N^i_j| > 1$, add pair-wise bi-directed arcs among elements of $N^i_j$.
2. Remove $\mathcal{M}^i$ from $\mathcal{D}^i$ to derive $\mathcal{R}^i$ (solving) and remove variables of $\mathcal{M}^i$ from $\mathcal{R}^i$ to derive $\mathcal{D}$ (substituting).
3. Let $i := i + 1$ and $\mathcal{D}^i := \mathcal{D}$.

**if** $\mathcal{D}^i$ is not empty

   **for each** remaining equation $e_k$ in $\mathcal{D}^i$, where $1 \leq k \leq |\mathcal{D}^i|$
   1. Create nodes $N_{e_k}$ for the set of variables, $V_{e_k}$, in $e_k$.
   2. Add arcs from nodes representing $OV_{e_k} \setminus V_{e_k}$ to $N_{e_k}$.
   3. Add pair-wise undirected arcs between nodes $N_{e_k}$.

Figure 2: Extended Causal Ordering Algorithm

---

model construction. Unlike the original causal ordering algorithm, each variable in the system is represented as a separate node so that the model builder can access and manipulate it directly. Directed arcs depict the causal relations among variables. In addition to these, our algorithm explicates the causal relations encoded in the under-constrained system. Bi-directed arcs denote feedback mechanisms in strongly-coupled subsets. User can visualize the effect of breaking the feedback system by manipulating one of variables connected by the bi-directed arc. Undirected arcs visually express relevant but undetermined causal relations among variables so that model builder can focus on clarifying the mechanisms governing these variables and complete the model.

**Example:** Suppose the budget planning officer wants to extend the StudentFacultyRatio model to take the average class size into account. She adds the structural equations $f_7 : CS = (NS * CL)/(NF * TL)$ and $f_8 : CL = 15$ to describe the relations among *ClassSize* (*CS*), *ClassLoad* (*CL*), *TeachingLoad* (*TL*), *NS* and *NF*. The structure matrix of the extended model is as follows:

|       | NS | NF | SFR | CS | CL | TL |
|-------|----|----|-----|----|----|----|
| $f_1$ | x  | 0  | 0   | 0  | 0  | 0  |
| $f_2$ | 0  | x  | 0   | 0  | 0  | 0  |
| $f_3$ | x  | x  | x   | 0  | 0  | 0  |
| $f_7$ | x  | x  | 0   | x  | x  | x  |
| $f_8$ | 0  | 0  | 0   | 0  | x  | 0  |

After applying the extended causal ordering algorithm to the system, she obtains the following under-constrained causal graph.

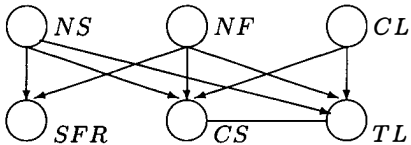

From the under-constrained causal graph, she can read off the current stage of problem formulation as follows: *StudentFacultyRatio* is determined by *NumberOfStudents* and *NumberOfFaculty*; currently both *ClassSize* and *TeachingLoad* depend on *NumberOfStudents*, *NumberOfFaculty*, and *ClassLoad*, but the relation between *ClassSize* and *TeachingLoad* is not yet determined, which is the consequence of the fact that the system is still under-constrained. □

### 3.3 Modeling Process

The modeling process starts with an initial focus, which is normally, in the spirit of value-focused thinking (Keeney 1994), the value variable. Users can also start with other focus variables, for example decision, observation, and whatever else is relevant or important a-priori. With the assistant interface, users can interactively browse the mechanisms related to their focus variables, select those that best depict the problem at hand, merge them, or specify exogenous variables to set the boundary of the system. However, we suggest the users to focus on one variable and add relevant mechanisms one at a time as the model evolves, since it resembles the action of focusing on a variable of interest, explaining or observing it in terms of its underlying mechanism. The user repeats the process iteratively until the model is requisite. In other words, users make decisions on the level of granularity and when to stop with the model building process. The system only plays the passive role of an assistant: suggesting mechanisms to choose from, indicating the possible mechanisms to merge, and denoting the manipulatable variables.

Normally a model evolves from an under-constrained system to a self-contained system. Designating manipulatable variables as exogenous helps in obtaining a self-contained system, i.e., orienting all arcs in the model graph. If the user assigns a potential policy variable, a manipulatable variable that is endogenous in a self-contained system, as exogenous, the whole model becomes over-constrained, because the number of equations is greater than the number of variables. We allow a model to be under-constrained or self-contained at any stage of the model development in *ImaGeNIe*, but we disallow a model to be over-constrained. When a model becomes over-constrained, the system pops up a list of mechanisms that are currently in the model and asks users to release one of them in order to change the system into a self-contained or an under-constrained system.

## 4 EXAMPLE MODEL BUILDING SESSION

We continue on extending our simple model to demonstrate how to interact with *ImaGeNIe* to build a simplified university budget model from UPBPM knowledge base encoded in *ImaGeNIe*. Suppose the officer has designated $TL$ variable as exogenous with equation $f_9 : TL = 6$. Figure 3 shows *ImaGeNIe* interface with the *navigation tree* of the knowledge base and the model we have built so far in the *workspace*.

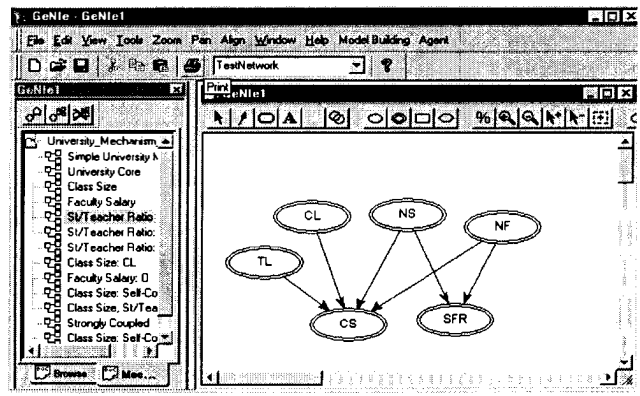

Figure 3: *ImaGeNIe* Interface: Navigation Tree and the Graphical Model Including Equations: $f_1$, $f_2$, $f_3$, $f_7$, $f_8$, and $f_9$.

Suppose she would like to plan the expenses related to faculty salary. She may use the *navigation tree* to locate mechanisms for *faculty salary*. Suppose she identifies the mechanism $f_{10} : FS = (OI + TA * NS)/(NF * (1 + O))$ that describes the interactions among variables: *FacultySalary* (*FS*), *OtherIncome* (*OI*), *TuitionAmount* (*TA*), *Overhead* (*O*), *NS* and *NF*. She drags it into the *workspace*. In order to maintain the unique variable identifiers in the model, *ImaGeNIe* automatically renames the $NS$ and $NF$ into $NS0$ and $NF0$. The extended causal ordering algorithm generates the graph shown in Figure 4.

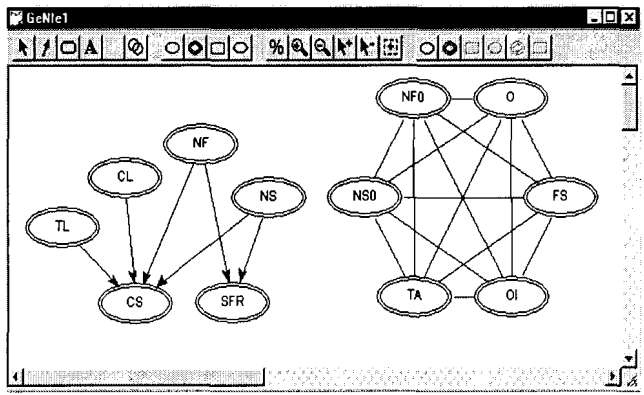

Figure 4: Model builder selects and drags $f_{10}$ into workspace; the extended causal ordering algorithm generates a corresponding graph.

She can then integrate the added mechanism with the model by merging $NS$ to $NS0$ and $NF$ to $NF0$. (See Figure 5).

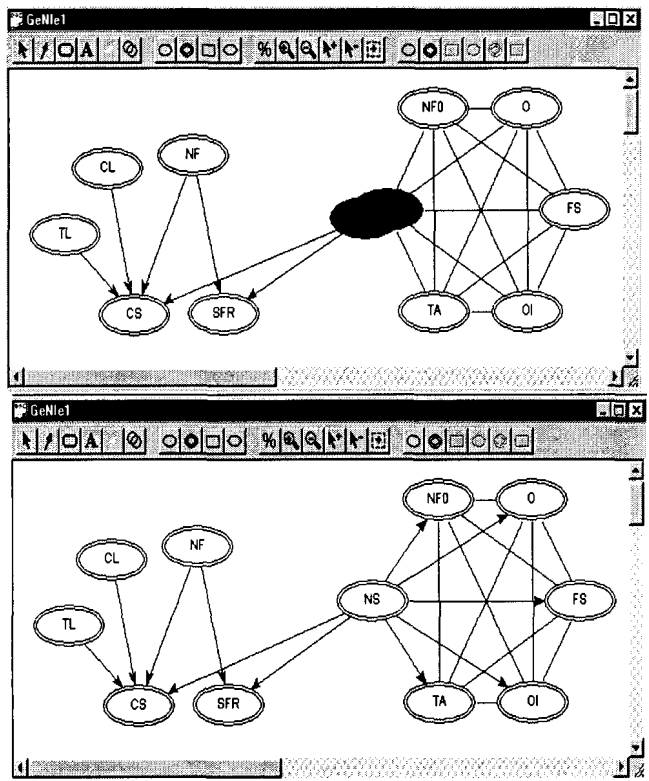

Figure 5: Model builder performs the merge operations for $NS$ (top). The causal ordering generates the corresponding graph (bottom).

She then makes $TA$, $O$, and $OI$ exogenous by assigning equations: $f_{11} : TA = 1,200$, $f_{12} : O = 0.48$, and $f_{13} : OI = 30,000,000$ and obtains a complete model that describes the dependence relations among those variables of interests (See Figure 6 top). She can now read off the following dependency relations from the complete model:

- Faculty salary is determined by the number of students, the number of faculty, tuition amount, other income, and overhead.
- Student-faculty ratio is determined by the number of students and the number of faculty.
- Class size is determined by the number of students, the number of faculty, class load, and teaching load.

After inspecting the current self-contained model, she would like to analyze the model under the condition that the average class size is fixed at 15 students per class. She makes the variable $CS$ exogenous by specifying a value assignment equation as $f_{14} : CS = 15$. Consequently the original self-contained model will become over-constrained. *ImaGeNIe* will ask her to release one of the equations (Figure 6 top). Suppose that she chooses to release the value assignment equation for the variable $TL$. The resulting graph generated by the causal ordering is shown in Figure 6 bottom.

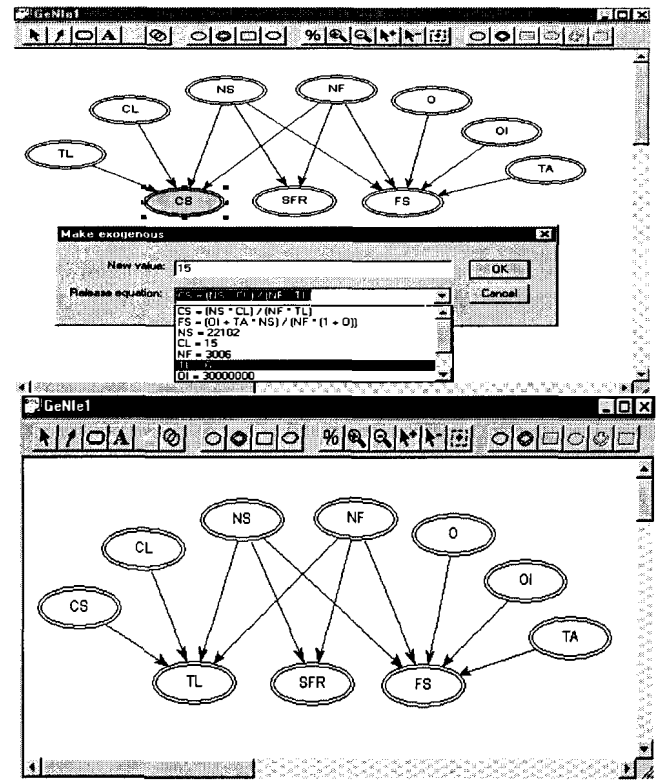

Figure 6: A change in structure on a self-contained model. The user manipulates $CS$ by setting $f_{14} : CS = 15$ and releasing $f_9 : TL = 6$ (top). The causal ordering generates the corresponding graph (bottom).

Now, she can read off the local effect of her change on the system from the causal graph: teaching load is determined by the number of students, the number of faculty, class load, and class size.

*ImaGeNIe* also supports model builder in extracting reusable mechanisms from the *workspace* into the knowledge base. Model builder simply selects the nodes of interest and drags them into the destination branch in the *navigation tree*. Due to space limitations, we are omitting this example.

## 5  DISCUSSION AND FUTURE WORK

Support for building model structure is one of the best ways of improving the quality of advice based on decision-theoretic models. While existing approaches focus on automatic model construction either from knowledge base or directly from data, our approach favors a closely-coupled loop between the system and its user. This is based on our belief that human judgement with respect to relevance, model size, completeness, and granularity is more reliable. Built on the assumption that under-constrained models reflect our problem recognition stages, *ImaGeNIe* assists users in encoding their conceptual problem framing in a causal graph generated by the extended causal ordering algorithm. Furthermore, *ImaGeNIe* provides users with the flexibility to choose building blocks from knowledge base to extend the model, to manipulate the variables in order to observe the effect of intervention (structure changes), and to extract reusable mechanisms from existing models to knowledge bases. The concept of causal mechanisms, on which *ImaGeNIe* relies, provides a general mean to accommodate different forms of knowledge description and makes knowledge acquisition task easier.

Recent research in applying the object-oriented framework to extend Bayesian networks for modeling complex domains (Koller & Pfeffer 1997; Laskey & Mahoney 1997; Pfeffer *et al.* 1999) is closely related to our work. Each of these approaches organizes domain knowledge into a hierarchical system. In Object-Oriented Bayesian Network (OOBN), the domain knowledge is structured explicitly as class-hierarchy for the type system and as object-hierarchy for the real model. In our framework, we do not impose any constraint on how users should organize their domain knowledge in the knowledge base. In the future, we would like to explore the semantics for combining type system with causal mechanisms so that our knowledge base can efficiently store the domain knowledge and be effectively used by users. As for the constructed models, *ImaGeNIe* provides submodels to group nodes into a graphical organization unit for the sake of succinct presentation, but there is no special semantic meaning attached to submodels in terms of inference. We plan to impose d-sepset (Xiang, Poole, & Beddoes 1993) constraint on submodels composition such that each submodel has well defined I/O sets to resemble object hierarchy in OOBN.

Once the model structures generated from our framework are associated with variable ranges and their numerical parameters such as explicit equations or conditional probability tables (CPTs), manipulation on the model may invalidate these numerical parameters. Druzdzel and van Leijen (2000) have shown the special conditions under which the CPTs in Bayesian networks can be reversed under manipulation. As for the explicit equations, *ImaGeNIe* tries to solve the manipulated system symbolically if there exists a solution. We would like to further explore conditions under which we can derive the numerical parameters from the mixture models after manipulation.

*ImaGeNIe* provides a flexible interactive model building environment for users to build models in causal form with as much system assistance as possible but without giving up their control over the model building process. We believe our efforts in incorporating causality as a heuristic in aiding model building and knowledge acquisition is an important extension to the existing approaches.

### Acknowledgments

This research was supported by the Air Force Office of Scientific Research, grants F49620-97-1-0225 and F49620-00-1-0112, by the National Science Foundation under Faculty Early Career Development (CAREER) Program, grant IRI-9624629, and by a strategic research grant number RP960351 from the National Science and Technology Board and the Ministry of Education in Singapore. We thank anonymous reviewers for suggestions improving the clarity of the paper. SMILE and GeNIe are available at http://www2.sis.pitt.edu/~genie.